\documentclass{article}

\usepackage{microtype}
\usepackage{graphicx}
\usepackage{subcaption}
\usepackage{booktabs} 

\usepackage{hyperref}

\usepackage[preprint]{icml2026}

\usepackage{amsmath}
\usepackage{amssymb}
\usepackage{mathtools}
\usepackage{amsthm}

\usepackage{booktabs} 
\usepackage{multirow} 

\usepackage[capitalize,noabbrev]{cleveref}
\usepackage[inline]{enumitem}
\theoremstyle{plain}

\theoremstyle{definition}

\theoremstyle{remark}

\usepackage[textsize=tiny]{todonotes}

\icmltitlerunning{Submission and Formatting Instructions for ICML 2026}

\begin{document}

\twocolumn[
  \icmltitle{Conditional Inverse Learning of Time-Varying Reproduction Numbers Inference}



  \icmlsetsymbol{equal}{*}

  \begin{icmlauthorlist}
    \icmlauthor{Lanlan Yu}{sch,comp}
    \icmlauthor{Quan-Hui Liu}{sch,comp}
    \icmlauthor{Haoyue Zheng}{sch,comp}
    \icmlauthor{Xinfu Yang}{sch,comp}
  \end{icmlauthorlist}

  \icmlaffiliation{sch}{College of Computer Science, Sichuan University, Chengdu, China}
  \icmlaffiliation{comp}{Engineering Research Center of Machine Learning and Industry Intelligence, Ministry of Education, Sichuan University, Chengdu, China}

  \icmlcorrespondingauthor{Quan-Hui Liu}{quanhuiliu@scu.edu.cn}

  \icmlkeywords{Machine Learning, ICML}

  \vskip 0.3in
]



\printAffiliationsAndNotice{}  

\begin{abstract}
    Estimating time-varying reproduction numbers from epidemic incidence data is a central task in infectious disease surveillance, yet it poses an inherently ill-posed inverse problem. Existing approaches often rely on strong structural assumptions derived from epidemiological models, which can limit their ability to adapt to non-stationary transmission dynamics induced by interventions or behavioral changes, leading to delayed detection of regime shifts and degraded estimation accuracy. In this work, we propose a Conditional Inverse Reproduction Learning framework (CIRL) that addresses the inverse problem by learning a {conditional mapping} from historical incidence patterns and explicit time information to latent reproduction numbers. Rather than imposing strongly enforced parametric constraints, CIRL softly integrates epidemiological structure with flexible likelihood-based statistical modeling, using the renewal equation as a forward operator to enforce dynamical consistency. The resulting framework combines epidemiologically grounded constraints with data-driven temporal representations, producing reproduction number estimates that are robust to observation noise while remaining responsive to abrupt transmission changes and zero-inflated incidence observations. Experiments on synthetic epidemics with controlled regime changes and real-world SARS and COVID-19 data demonstrate the effectiveness of the proposed approach.
\end{abstract}

\section{Introduction}
The estimation of the time-varying reproduction number, $R_t$, is fundamental to characterizing the dynamics of infectious disease outbreaks and guiding public health interventions~\cite{wallinga2004different,fraser2011influenza,gostic2020practical,trevisin2023spatially}. Values of $R_t$ above or below the critical threshold of 1 indicate expanding or declining epidemics, respectively~\cite{wallinga2007generation,liu2020covid}. From a mechanistic perspective, the transmission dynamics is governed by the renewal process~\cite{kermack1927contribution}. The expected incidence at time $t$, denoted as $\mathbb{E}[I_t]$, is modeled as the product of the instantaneous reproduction number and the total infectiousness of the population~\cite{cori2013new}:
\begin{equation}
\mathbb{E}[I_t]
= R_t \sum_{\tau=1}^{\infty} I_{t-\tau} \, w(\tau),
\label{eq:renewal}
\end{equation}
where $I_t$ denotes the number of newly infected individuals in discrete time $t$, the time-varying reproduction number $R_t$ denotes the ratio of the number of new infections generated at time step $t$, and $w(\tau)$ represents the serial (or generation) interval distribution, satisfying $\sum_{\tau=1}^{\infty} w(\tau) = 1$. This equation defines a forward operator mapping latent transmission dynamics to the observed cases~\cite{dai2023new}.

Since $R_t$ is unobserved, inferring it from noisy and potentially delayed incidence data $\mathcal{I}= \{I_t\}_{t=1}^T$ constitutes a classic \emph{ill-posed inverse problem} \cite{benning2018modern, parag2021improved}. Multiple latent trajectories may explain the same observations, and small perturbations in the data can induce large variations in inferred dynamics. It is therefore necessary to constrain the solution space in order to recover meaningful latent transmission trajectories.
\cite{wallinga2004different,cori2013new,abbott2020estimating}.

To mitigate this ill-posedness, standard approaches typically impose \emph{hard constraints}—rigid structural assumptions designed to force a unique solution~\cite{wallinga2004different,cori2013new,abbott2020estimating,song2022estimating}. 
However, recent theoretical analyses indicate that hard-constrained or purely deterministic $R_t$ estimation can be problematic, since rigid assumptions are incompatible with non-stationary transmission dynamics~\cite{ghosh2023approximate}, especially in scenarios with early missing observations and sudden shifts driven by interventions~\cite{dai2023new,lau2021evaluating}.
For instance, the local stationarity assumption inherent in sliding-window methods (e.g., EpiEstim) fundamentally contradicts the instantaneous changes of intervention-induced discontinuities~\cite{cori2013new,parag2021improved}. Similarly, neural methods rooted in differential equations often implicitly enforce exponential generation intervals (memoryless dynamics)~\cite{song2022estimating}, diverging from the non-Markovian and context-dependent nature of real transmission processes\cite{pang2025time}.
When transmission dynamics change abruptly, such structural mismatches can cause model to conflate transient reporting artifacts with genuine epidemiological signals, leading to detection lags and smoothed trajectories~\cite{nash2022real}. This failure mode mirrors convergence pathologies frequently observed in scientific machine learning when physical laws are enforced on imperfectly observed systems~\cite{karniadakis2021physics, lu2021physics}.

Second, deterministic frameworks are ill-equipped to handle stochasticity arising from missing observations, leading to poorly calibrated uncertainty~\cite{wang2025bayesian}. 
Observation processes in realistic surveillance are distorted by structural zeros (e.g., due to limited diagnostic capacity) and stochastic delays, which rigid or purely deterministic optimization tends to misinterpret as meaningful transmission changes~\cite{parag2021improved, abbott2020estimating, lison2024generative}. 
While probabilistic frameworks like EpiNow2~\cite{abbott2020estimating} introduce smoothness priors to mitigate noise, they typically lack explicit mechanisms to model these systematic reporting failures.
This limitation results in poorly calibrated uncertainty quantification, particularly in low-incidence regimes where the model produces dangerously narrow confidence intervals around potentially biased estimates~\cite{lison2024generative,hettinger2025refining}.

In this work, we advocate a shift from rigid structural assumptions toward a \emph{conditional inverse learning}~\cite{arridge2019solving,ongie2020deep} perspective for estimating time-varying reproduction numbers, which we term \textbf{Conditional Inverse Reproduction Learning (CIRL)}.
Instead of imposing piecewise-constant dynamics or window-based smoothing constraints on $R_t$, we reformulate the problem as learning a flexible conditional inverse mapping, $R_t = f_\theta(t, \mathcal{H}_t)$, conditioned on the recent explicit time coordinate $t$ and epidemiological history $\mathcal{H}_t$. 
Under this formulation, the admissible solution space is implicitly constrained by the conditioning structure, rather than by hard-coded temporal assumptions.

CIRL consists of two key components.
(i) A \emph{Conditional Inverse Mapping Network} that parameterizes $f_\theta(\cdot)$ using a neural architecture conditioned on historical observations and time.
(ii) A \emph{Statistical Observation and Consistency Module}, which replaces deterministic fitting targets with a probabilistic objective to accommodate the heterogeneity and sparsity of real-world surveillance data, including over-dispersion, reporting artifacts, and early missing observations.
This soft consistency constraint suppresses noise-induced high-frequency oscillations while retaining sensitivity to abrupt transmission changes when supported by sufficient evidence in the conditioning history.
Since conditional inverse mapping depends only on past incidence, the framework can be applicable in prospective estimation settings, while naturally enabling short-term forward simulation as an auxiliary check of epidemiological consistency. 

Our key contributions are summarized as follows:
\begin{itemize}
    \item We reformulate $R_t$ estimation as a \emph{probabilistic conditional inverse problem} under the renewal equation, avoiding hard-coded temporal assumptions on $R_t$.
    \item We introduce a \emph{likelihood-based consistency} formulation that replaces deterministic fitting targets, enabling principled handling of heterogeneous and sparse surveillance data.
    \item We empirically validate the proposed framework on synthetic epidemics and real epidemic data, demonstrating robust inverse inference under under-reporting, missing observations, and abrupt transmission changes.
\end{itemize}

\section{Problem Formulation}
\label{sec:problem_formulation}
The time-varying reproduction number ($R_t$) quantifies the average number of secondary infections generated by infectious individuals at time $t$. 
We formulate the epidemic dynamics as a discrete-time process with observed incidence $\mathcal{I} = \{I_t\}_{t=1}^T$, where $I_t \in \mathbb{N}$ denotes the number of newly reported cases at time $t$. The goal is to recover the underlying time-varying reproduction numbers, denoted as $\mathcal{R} = \{R_t\}$, which are treated as latent, time-varying variables inferred from the observables $\mathcal{I}$.

\paragraph{Forward Renewal Equation.}
The relationship between latent transmission trajectories $\mathcal{R}$ and observed cases $\mathcal{I}$ is governed by the renewal equation:
\begin{equation}
    I_t \sim \mathcal{D}(\lambda_t),
    \label{eq:renewal_generation}
\end{equation}
where $\mathcal{D}$ denotes a discrete count distribution with expectation $\lambda_t$. The term $\lambda_t$ denotes the conditional expectation:
\begin{equation}
\lambda_t = R_t \sum_{\tau=1}^{t-1} I_{t-\tau} w_\tau,
\label{eq:renewal_lam}
\end{equation}
where $w_\tau$ is the empirical generation interval distribution with $\sum_{\tau} w_\tau = 1$. This defines a forward mapping $\mathcal{F} : \mathcal{R} \to \mathcal{I}$, which is both many-to-one and one-to-many. Given the renewal intensity $\lambda_t$, the observed incidence $I_t$ is modeled as a noisy count process centered at $\lambda_t$, potentially subject to reporting noise and over-dispersion.

\paragraph{Conditional Inverse Mapping.}
Given observed incidence $\mathcal{I}$ and a known generation interval distribution $\{w_\tau\}$, estimating $R_t$ corresponds to the inverse recovery under the renewal equation. This problem is inherently ill-posed, since multiple $R_t$ trajectories can produce identical incidence sequences. 
To address this ambiguity, our objective is to learn a conditional inverse mapping $\mathcal{F}^{-1}:\mathcal{I} \to \mathcal{R}$ that infers latent $R_t$ sequences consistent with the observed epidemic dynamics while leveraging past incidence as conditioning information, as follows:
\begin{equation}
\mathcal{F}^{-1}: R_t = f_\theta(t, \mathcal{H}_t),
\label{eq:inverse_mapping}
\end{equation}
where $f_\theta$ is a parameterized inverse mapping learned from the data, $t$ denotes the explicit time coordinate, and $\mathcal{H}_t = \{I_{t-L}, \dots, I_{t-1}\}$ defines a historical context (receptive field) of size $L$.Accordingly, Eq.~\eqref{eq:inverse_mapping} applies only to $t > L$, as the initial window lacks sufficient historical context to infer $R_t$.
It is worth emphasizing that our conditioning horizon $L$ solely defines the temporal context available to the conditional inverse mapping, and does not impose any piecewise-constant, local smoothness, or stability assumptions on $R_t$. 
Under this formulation, $R_t$ inference is formulated as a conditional inverse learning problem grounded in epidemiological renewal dynamics.

\begin{figure*}[ht]
    \centering
    \includegraphics[width=0.95\linewidth]{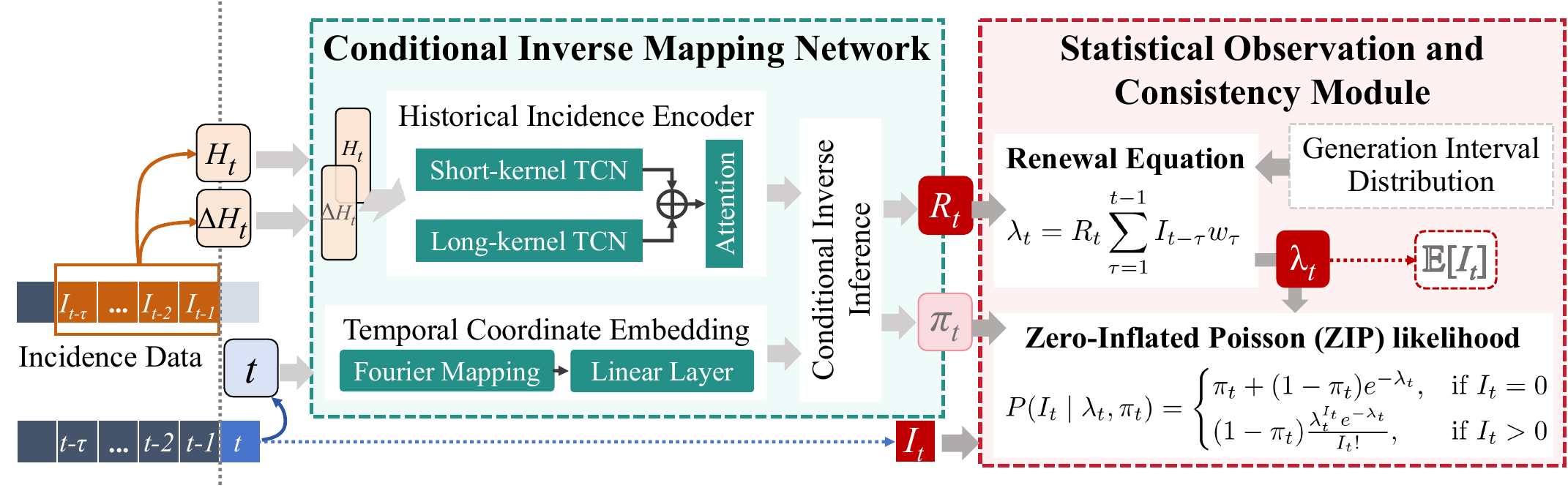}
    \caption{Overview of {Conditional Inverse Reproduction Learning} (CIRL) framework}
    \label{fig:overview-model}
\end{figure*}

\section{Method}
We propose a \emph{Conditional Inverse Reproduction Learning} (CIRL) framework that learns a parameterized conditional inverse mapping $f_\theta$ associated with the renewal operator $\mathcal{F}$ to infer $R_t$ from observed incidence data. 
As illustrated in Figure~\ref{fig:overview-model}, $f_\theta$ is conditioned on the retrospective epidemic history $\mathcal{H}_t$ and the temporal coordinate $t$, and outputs latent transmission dynamics $\mathcal{R}$, from which the expected incidence $\mathbb{E}[I_t]$ is obtained via the renewal equation (Eq.~\eqref{eq:renewal}).

Based on the architecture of $f_\theta$ and the strategy for learning its parameters, the CIRL framework is organized into two complementary components:
(i) the \emph{Conditional Inverse Mapping Network}, which maps historical incidence and temporal context to latent transmission dynamics, and
(ii) the \emph{Statistical Observation and Consistency Module}, which enforces consistency with observed incidence through a probabilistic renewal-based formulation.
Together, these components enable flexible, data-driven inference of $R_t$ while respecting epidemiological dynamics.

\subsection{Conditional Inverse Mapping Network}
We instantiate $f_{\theta}$ using a modular neural architecture designed to capture non-stationarity, temporal dependence, and multi-scale epidemic dynamics.
The conditional inverse mapping $f_\theta (t_c, \mathcal{H}_{t_c})$ is implemented as a modular neural architecture that decomposes the inference task into distinct functional components. Specifically, the architecture consists of: 
(i) a temporal embedding module that encodes explicit time information,
(ii) a historical incidence encoder that extracts multi-scale features from past observations,
(iii) a fusion module that produces a latent reproduction number $R_{t_c}$ at time $t_c$.

\paragraph{Temporal Coordinate Embedding (TCE).}\label{subsec:time_embedding}
The explicit time index $t_c$ is embedded via a coordinate-based encoding,
\begin{equation}
\mathbf{e}_{t_c} = \phi_{\text{time}}(\boldsymbol{\phi}_\text{FF}(t_c)),
\end{equation}
where $\phi_{\text{time}}(\cdot)$ is a learnable embedding function implemented as a multilayer perceptron (MLP). Here, $\boldsymbol{\phi}_\text{FF}(t_c)$ is a Fourier feature mapping about time $t_c$:
\begin{equation}
\boldsymbol{\phi}_\text{FF}(t_c) =
\begin{bmatrix}
\cos(\alpha \pi t_c) \\
\sin(\alpha \pi t_c)
\end{bmatrix},
\end{equation}
where $\alpha$ is a user-defined, non-trainable frequency vector. This mapping addresses spectral bias, enabling the model to efficiently capture both high-frequency variations and long-range temporal dependencies, going beyond local patterns to model global temporal effects and long-term non-stationarity~\cite{ning2023epi}.

\paragraph{Historical Incidence Encoder (HIE).}\label{subsec:historical_embedding}
To extract informative representations from the historical incidence context $\mathcal{H}_{t_c}$, we employ a sequence encoder:
\begin{equation}
\mathbf{h}_{t_c} = \phi_{\mathrm{hist}}\big([\mathcal{H}_{t_c}; \Delta \mathcal{H}_{t_c}]\big),
\end{equation}
where $\phi_{\mathrm{hist}}$ is designed to capture epidemic dynamics at multiple temporal scales, via multi-scale Temporal Convolutions Network (TCN)~\cite{szegedy2015going,bai2018empirical} with different receptive fields. 
In particular, to augment the representational capacity of the temporal encoder, we incorporate the first-order temporal difference $\Delta \mathcal{H}_{t_c} \leftarrow \{\Delta I_t = I_t - I_{t-1}, t\leq t_c\}$ as a secondary input channel. This explicitly provides the model with instantaneous growth dynamics, enabling the TCN to effectively distinguish between different epidemic phases (e.g., peak deceleration vs. early-stage acceleration) even when absolute incidence magnitudes are comparable.

To capture diverse temporal dynamics, we use an Inception-style convolutional block with two parallel TCN branches employing different kernel sizes (e.g., $\mathcal{K}=\{3,7\}$). The short-term branch detects rapid fluctuations and localized outbreaks, while the long-term branch smooths noise and tracks stable trends. Both branches use dilated causal convolutions for a large receptive field while preserving temporal causality, generating multi-scale feature maps $\mathbf{F}^{short}_{t_c}$ and $\mathbf{F}^{long}_{t_c}$. The concatenated multi-scale features are projected into a sequence of embeddings, which are then processed by a Transformer Encoder~\cite{vaswani2017attention}:
\begin{equation}
    \mathbf{h}_{t_c} = \mathrm{Attn}\big([\mathbf{F}^{short}_{t_c}; \mathbf{F}^{long}_{t_c}]\big).
\end{equation}
By leveraging the self-attention mechanism, the model learns to selectively weight specific historical time points that are most informative for the current $R_{t_c}$ state, effectively capturing non-local temporal dependencies that are often missed by purely convolutional filters.

\paragraph{Conditional Inverse Inference of $R_t$}\label{subsec:fusion_encoder}
The temporal coordinate  embedding $\mathbf{e}_{t_c}$ and historical incidence feature $\mathbf{h}_{t_c}$ are integrated through a nonlinear fusion operator:
\begin{equation}
\mathbf{z}_{t_c} = \phi_{\mathrm{fusion}}(\mathbf{e}_{t_c}, \mathbf{h}_{t_c}),
\end{equation}
where $\phi_{\mathrm{fusion}}(\cdot)$ is implemented as feature concatenation followed by a Cross-Attention module~\cite{vaswani2017attention} to capture interactions between global temporal context and local historical dynamics. 

The latent reproduction number is then obtained as
\begin{equation}
R_{t_c} = f_{\theta}(t_c,\mathcal{H}_{t_c}) = \phi_{\mathrm{out}}(\mathbf{z}_{t_c};t_c,\mathcal{H}_{t_c},\theta),
\label{eq:conditional-inverse-mapping}
\end{equation}
where $\phi_{\mathrm{out}}(\cdot)$ is implemented as an MLP followed by a positivity-preserving activation function.
This design introduces a low-dimensional bottleneck between historical observations and transmission dynamics, which regularizes the inverse mapping and improves estimation stability.

\subsection{Statistical Observation and Consistency Module}
Given the conditional inverse mapping in Eq.~(\eqref{eq:conditional-inverse-mapping}), here, we describe how the parameters $\theta$ of $f_\theta(t_c, \mathcal{H}_{t_c})$ are learned from observed incidence using a physics-informed statistical objective, enabling accurate inference of the latent reproduction numbers.
To this end, the \emph{Statistical Observation and Consistency Module} defines a probabilistic, renewal-based likelihood that enforces coherence between the inferred $R_t$ trajectory and observed incidence, while robustly handling over-dispersion, reporting noise, and observation gaps via a zero-inflated objective. 
By optimizing the network parameters under this objective, the framework learns a conditional inverse mapping that is both epidemiologically consistent and flexible enough to capture complex temporal transmission patterns.

\paragraph{Renewal-based Zero-Inflated Poisson Observation Model.}
As defined Eq.~\eqref{eq:renewal_lam} in the problem formulation, the renewal equation acts as a forward operator that maps the latent reproduction number $R_t$ and historical incidence to the expected number of new cases $\lambda_t$.
This forward mapping enables us to enforce statistical consistency between the inferred transmission dynamics and the observed epidemic curve.
Rather than treating the renewal equation as a hard constraint, we incorporate it as a fully differentiable structural component within the learning objective. This design facilitates gradient flow through epidemiological laws, guiding the model toward physically plausible solutions without compromising optimization flexibility.
To explicitly account for the stochasticity, reporting noise, and zero-inflation characteristic of surveillance data, we model the observed incidence $I_t$ using a robust probabilistic count distribution conditioned on the renewal expectation $\lambda_t$ (Eq.~\eqref{eq:renewal_lam}).

Real-world surveillance data is characterized by two distinct types of noise: over-dispersion (variance exceeding the mean) and zero-inflation (excess zeros due to under-reporting or low transmission periods). Standard Poisson objectives fail to capture these dynamics. Therefore, we model the observed incidence $I_{t}$ using a Zero-Inflated Poisson (ZIP) distribution~\cite{lambert1992zero}. The probability mass function is defined as:
\begin{equation}
    P_{ZIP}(I_t \mid \lambda_t, \pi_t) = 
    \begin{cases} 
    \pi_t + (1-\pi_t) e^{-\lambda_t}, & \text{if } I_t = 0 \\
    (1-\pi_t) \frac{\lambda_t^{I_t} e^{-\lambda_t}}{I_t!}, & \text{if } I_t > 0 
    \end{cases}~.
\end{equation}
Notably, unlike standard ZIP models that assume a static $\pi$, we estimate $\pi_t$ as a time-varying latent parameter via a parallel network branch (activated by Sigmoid), allowing the model to dynamically account for structural reporting failures.
By optimizing the likelihood of the observed data under this distribution, our model creates a robust buffer against observation noise, preventing the "over-fitting to noise" problem common in deterministic inversion.
We formulate the inference of $R_t$ as a regularized maximum likelihood estimation problem, yielding the data fidelity term
\begin{equation}
\mathcal{L}_{\text{obs}} = -\log P_{ZIP}(I_t \mid \lambda_t, \pi_t),
\end{equation}
which encourages the inferred reproduction numbers to generate incidence trajectories that are statistically aligned with observations.

\paragraph{Temporal Smoothness Regularization of $R_t$.}
To mitigate the inherent ill-posedness of inferring $R_t$ from noisy incidence, we impose a smoothness regularization~\cite{rudin1992nonlinear} on the inferred reproduction number sequence.
Specifically, we penalize abrupt changes in $R_t$ by applying a robust loss to its first-order temporal differences:
\begin{equation}
\mathcal{L}_{\text{smooth}} = \sum_t \ell_{\text{Huber}}(R_t, R_{t-1}),
\end{equation}
where $\ell_{\text{Huber}}(\cdot)$ denotes the Huber loss~\cite{huber1992robust}, serving as a differentiable approximation of Total Variation (TV) regularization~\cite{rudin1992nonlinear}.
This formulation suppresses spurious high-frequency fluctuations while remaining tolerant to minor stochastic variations and preserving sensitivity to genuine regime shifts.

\paragraph{Overall Training Objective.}
The final learning objective integrates observation consistency with physics-informed regularization:
\begin{equation}
\mathcal{L} =
\mathcal{L}_{\text{obs}}
+ \omega \mathcal{L}_{\text{smooth}},
\end{equation}
where $\omega$ control the relative strength of temporal smoothness.
All components are fully differentiable, enabling end-to-end optimization of the conditional inverse operator via gradient-based methods.

\subsection{Inference and Usage}
After training, the learned inverse mapping $f_{\theta^(t, \mathcal{H}_t)}$ enables retrospective estimation of $\mathcal{R}$ for the training window and real-time inference for incoming data ($t \ge T$). Additionally, the corresponding incidence trajectories can be reconstructed by recursively applying the renewal equation (Eq.~\eqref{eq:renewal}) with the inferred rates.

\section{Experimental Evaluation}
\label{sec:experiments}
\subsection{Experimental Setup}
\paragraph{Datasets.} 
To evaluate the robustness and generalization capability of our framework, we conducted experiments on synthetic benchmarks and two diverse real-world epidemic scenarios.

\begin{itemize}
    \item Synthetic datasets were generated by sampling incidence counts from a Poisson distribution~\cite{dai2023new}, using a SARS-like generation interval (Mean=$8.0$, SD=$3.0$) under two $R_t$ profiles: single-step and double-step changes, with a low seed of 2 infected individuals ($I_0=2$).
    Furthermore, we applied a \emph{proportional zero-inflated mask} (pre-peak $p=0.3$, post-peak $p=0.05$) to simulate realistic surveillance evolution and stress-test the model against structural sparsity.
    
    \item Real-world datasets consisted of (i) SARS 2003 (Hong Kong)~\cite{cori2009temporal}, characterized by data sparsity and a long generation interval \citep{lipsitch2003transmission}, and (ii) COVID-19 (Ontario)~\cite{song2022estimating}, representing a dense, high-noise scenario with a short generation interval \citep{knight2020estimating}. 
\end{itemize}

\paragraph{Evaluation Metrics.}
Synthetic experiments are evaluated using RMSE as the primary accuracy metric, MAE as a robustness measure, and detection delay to quantify responsiveness to regime changes:
\begin{align}
\mathrm{RMSE} &= \sqrt{\frac{1}{T'} \sum_{t \in \mathcal{T}} \left( \hat{R}_t - R_t^{\star} \right)^2}, \\
\mathrm{MAE} &= \frac{1}{T'} \sum_{t \in \mathcal{T}} \left| \hat{R}_t - R_t^{\star} \right|,
\end{align}
where $\hat{R}_t$ denotes the inferred reproduction number, $\mathcal{T}$ is the set of valid time $t\ge L$, and $T' = |\mathcal{T}|$.

Beyond pointwise accuracy, we evaluated the responsiveness of inferred reproduction numbers to shifts in transmission dynamics.
For scenarios with known change points (e.g., abrupt or multiple regime shifts), we measure the detection delay as
\begin{equation}
\Delta_{\text{delay}} = \hat{t}_{\text{cp}} - t_{\text{cp}},
\end{equation}
where $t_{\text{cp}}$ denotes the first true change point, and $\hat{t}_{\text{cp}}$ is the first index where the inferred $\hat{R}_t$ crosses the epidemic threshold $R_t = 1$. To assess methods' ability to detect regime shifts, we report the Missed Detection Rate (MDR), defined as the fraction of simulations in which $\hat{R}_t$ fails to cross the threshold at the true change point.

To emulate the genesis of real-world epidemics, which typically originate from a few index cases, all simulations were initialized with a low seed of $I_0=2$. This realistic initialization naturally predisposes the system to stochastic extinctions and high variance. Therefore, we summarize performance metrics using the median and interquartile interval ($[Q_1, Q_3]$) rather than the mean and standard deviation.

For real epidemic data, where $R_t^{\star}$ is unobserved, we do not report regression-based accuracy metrics.
Instead, we assessed qualitative consistency, temporal plausibility, and forward validation through short-term incidence forecasting (10 days).
Specifically, we evaluate the predictive performance of the implied incidence $\hat{I}_{t+h}$ (Eq.~\eqref{eq:renewal}) using standard metrics (e.g., RMSE and MAE) over short horizons, serving as an auxiliary check on the inferred transmission dynamics.

\paragraph{Baselines.} To strictly evaluate the proposed framework, we compare it with three representative methods, categorized by their strategy to solve the ill-posed inverse problem of $R_t$ estimation:
\begin{itemize}
    \item EpiEstim~\cite{cori2013new} is a widely used Bayesian framework that estimates $R_t$ by locally inverting the renewal equation under a sliding window assumption. Within each window, $R_t$ is assumed to be constant and inferred using a Poisson likelihood with a Gamma prior.

    \item EpiNow2~\cite{abbott2020estimating} extends the renewal equation framework by incorporating reporting delays, observation noise, and time-varying reproduction numbers within a Bayesian state-space model.
    It models latent infections and $R_t$ jointly through hierarchical probabilistic inference, providing posterior uncertainty estimates.
    
    \item Deep Learning–based Estimation (UDENet)~\cite{song2022estimating} parameterizes $R_t$ using neural networks trained on simulated epidemic data. This model learns a direct mapping from the recent incidence to reproduction numbers without explicit enforcement of the renewal equation during inference.
\end{itemize}

All baseline methods are provided with the same incidence data and generation interval distribution.

\begin{table*}[t]
\caption{Quantitative comparison of different methods on synthetic benchmarks over 100 simulations. For continuous metrics (RMSE, MAE, and $\Delta_{\text{delay}}$), results are reported as Median [$Q_1, Q_3$]. Bold values indicate the best performance in each column, and underlined values indicate the second-best.}
\label{tab:abrupt-results}
\begin{center}
\begin{tiny}
\setlength{\tabcolsep}{2pt} 
\begin{tabular}{l rccc rccc rccc}
\toprule
\multirow{2}{*}{Method}  & \multicolumn{4}{c}{Single-step} & \multicolumn{4}{c}{Double-step} & \multicolumn{4}{c}{Double-step(masked)}\\
\cmidrule(r){2-5} \cmidrule(lr){6-9} \cmidrule(lr){10-13} 
& $\Delta_{\text{delay}}$ & MDR & RMSE & MAE & $\Delta_{\text{delay}}$ & MDR & RMSE & MAE & $\Delta_{\text{delay}}$ & MDR& RMSE & MAE  \\
\midrule
EpiEstim
& \underline{6.00}[5.00,6.00] & 0.00 & \textbf{0.36}[0.31,0.42] & \textbf{0.22}[0.19,0.26]
& 6.00[5.00,6.00] & 0.00 & \textbf{0.40}[0.37,0.49] & \textbf{0.27}[0.24,0.31] 
& \underline{6.00}[-9.00,12.00]& 0.00 & 0.83[0.57,1.76] & 0.50[0.38,0.74] 
\\

EpiNow2
& 8.00[7.00,11.00] & 0.00 & \underline{0.35}[0.32,0.39] & \textbf{0.22}[0.20,0.27] 
& 47.00[27.50,49.00] & 0.01 & \underline{0.45}[0.37,0.59] & 0.36[0.27,0.47]
& 52.00[50.00,55.00] & 0.03 & \underline{0.57}[0.52,0.62] & \underline{0.49}[0.45,0.52] 
\\

UDENet
& -39.00[-39.50,7.00] & 0.97 & 0.60[0.47,2.04] & 0.52[0.40,1.86] 
& -29.00[-29.20,-28.00] & 0.97 & 1.53[0.55,2.17] & 1.35[0.51,2.03] 
& $\sim$ & 1.00 & 0.61[0.55,1.93] & 0.56[0.51,1.75] 
\\

CIRL(ours)
& \textbf{1.00}[0.00,1.00] & 0.01 & 0.38[0.32, 0.48] & \underline{0.28}[0.23,0.33] 
& \textbf{1.00}[0.00,2.00] & 0.01 & 0.48[0.36,0.56] & \underline{0.33}[0.26,0.39] 
& \textbf{5.00}[2.00,9.00] & 0.02 & \textbf{0.51}[0.44,0.58] & \textbf{0.36}[0.31,0.42] 
\\

\bottomrule
\end{tabular}
\end{tiny}
\end{center}
\vskip -0.1in
\end{table*}

\subsection{Performance on Synthetic Benchmarks}
To systematically evaluate the proposed method under controlled transmission dynamics, we design a set of synthetic epidemic benchmarks with known time-varying reproduction numbers. Specifically, we consider abrupt regime-change scenarios with single or double-step changes and zero-inflated observation
regimes, for which we generate 100 stochastic simulations with identical parameters and assess estimation accuracy (RMSE and MAE), regime shift detection rate (MDR), and shift detection delay ($\Delta_{\text{delay}}$). Supplementary analysis of incidence reconstruction is provided in Appendix~\ref{app:incidence-reconstruction}.

\begin{figure}[ht]
    \centering
        \centering
        \includegraphics[trim=0pt 0pt 0pt 17pt, clip, width=0.78\linewidth]{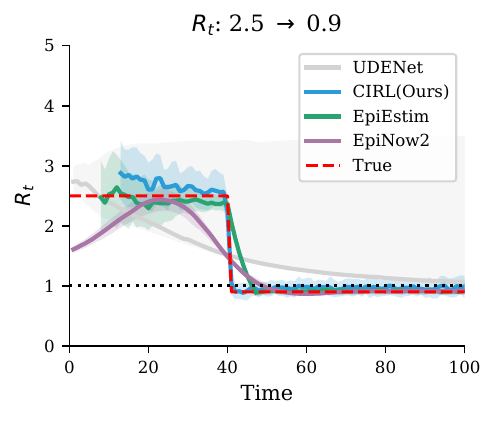}

    \caption{Median and interquartile range of the inferred $R_t$ over 100 simulations. Solid lines denote the median estimate across 100 simulations, while shaded regions indicate the interquartile range (25\%–75\%). (a) $R_t$ estimation across single-step change scenarios, while (b) estimation on double-step change scenarios. .
}
    \label{fig:single-results}
\end{figure}

\paragraph{Single regime-change scenario.} 
We examine stochastic settings characterized by single-step $R_t$ shifts, designed to emulate the abrupt discontinuities in transmission dynamics triggered by stringent non-pharmaceutical interventions (NPIs), such as city-wide lockdowns.
Figure~\ref{fig:single-results} reveals distinct failure modes inherent to the baseline methods, traceable to their rigid structural priors.
UDENet fails to detect shifts (MDR=97\% in Table~\ref{tab:abrupt-results}) due to its unrealistic memoryless generation interval formulation. 
EpiEstim is fundamentally limited by its rigid block-constant assumption, which presumes $R_t$ remains strictly constant within the sliding window (e.g., $\tau=7$). This assumption acts as a temporal smoothing filter, rendering the method incapable of resolving instantaneous shifts and leading to a significant median delay of 6 time steps in Table~\ref{tab:abrupt-results}.
Similarly, EpiNow2, due to its reliance on Gaussian Process smoothing priors to enforce temporal continuity, tends to over-regularize abrupt changes, producing overly smooth trajectories.
Conversely, our data-driven approach avoids these structural biases, achieving near-instantaneous detection (median delay = 1 step), while maintaining competitive accuracy (RMSE/MAE within 0.03/0.06 of the best baseline). In this case, we also note a single failure instance (MDR=$0.01$) where CIRL interpreted the shift as noise. In greater detail, the post-hoc analysis of Appendix~\ref{app:failure_analysis} reveals that this specific failure occurred in an extreme 'low-information' regime: the simulation was characterized by a zero-proportion exceeding $50\%$ and a peak incidence of only $4$.

\begin{figure}[ht]
    \centering
    \begin{subfigure}[b]{0.48\linewidth}
        \centering
        \includegraphics[trim=0pt 0pt 0pt 17pt, clip, width=\linewidth]{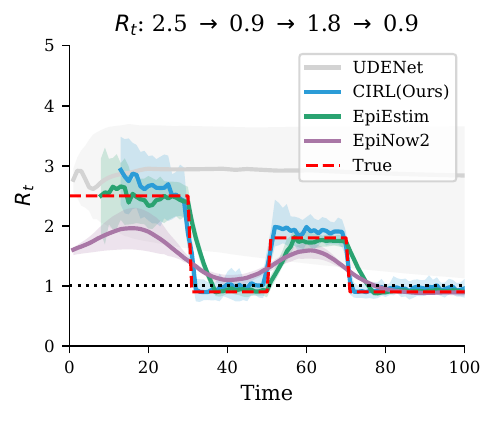}
        \caption{Poisson-generated incidence} 
        \label{fig:double}
    \end{subfigure}
    \hfill 
    \begin{subfigure}[b]{0.48\linewidth}
        \centering
        \includegraphics[trim=0pt 0pt 0pt 17pt, clip, width=\linewidth]{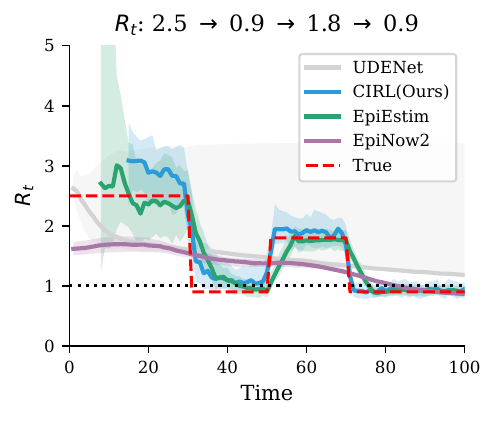}
        \caption{Poisson incidence with mask} 
        \label{fig:double-zip}
    \end{subfigure}

    \caption{Estimated reproduction number trajectories under different observation noise conditions. Both panels show estimates obtained from a single synthetic epidemic with smooth oscillation dynamics. (a) corresponds to incidence generated from a Poisson process, while (b) illustrates the same epidemic after introducing proportional zero-inflated mask in the observations. The comparison highlights the robustness of the inferred transmission dynamics to observation sparsity.
}
    \label{fig:double-results}
\end{figure}

\paragraph{Double regime-change scenario.} 
We further extend the evaluation to a double-step regime, designed to emulate the dynamics of stringent intervention (precipitating a sharp drop in $R_t$) followed by policy relaxation (leading to viral resurgence).
The results corroborate our single-step analysis, as shown in Figure~\ref{fig:double}. In this complex setting, UDENet remains effectively blind to these changes (MDR=97\%), with its median trajectory failing to register the shifts entirely. Among the other baselines, EpiNow2 consistently over-smooths transitions, while EpiEstim exhibits a systematic detection lag at every change point. 
In stark contrast, our method captures every structural shift with a median $\Delta_{delay}$ of only 1 time step. Although this responsiveness comes with a negligible trade-off in median precision (a gap of $<0.08$ in RMSE compared to EpiEstim in Table~\ref{tab:abrupt-results}), our method provides significantly higher estimation certainty, evidenced by a tighter Interquartile Range (IQR) compared to EpiNow2.

\paragraph{Zero-Inflated Observation Regimes.} 
Real-world surveillance, particularly in early outbreak stages, is often plagued by under-reporting due to limited diagnostic capacity and asymptomatic transmission. 
To emulate this, we introduce a state-dependent sparsity mechanism into the double-step scenario, applying a proportional zero-inflate mask.
Under this stress test, UDENet collapses (MDR=100\%), and EpiNow2 fails to identify clear change points. EpiEstim confirms its structural limitation, lagging by a median of 6 steps.
Notably, the high density of zeros in the early phase creates an identifiability bottleneck for the initial $R_t \to 0$ transition, affecting both our method and EpiEstim, shown in Figure~\ref{fig:double-zip}. 
However, whereas EpiEstim produces highly unstable estimates (large IQR for RMSE in Table~\ref{tab:abrupt-results}), our framework demonstrates remarkable resilience. 
This robustness arises from our Physics-Informed Statistical Learning module, which mitigates the influence of observation artifacts and enables more reliable learning of the latent transmission dynamics, resulting in consistently strong performance across RMSE, MAE, and $\Delta_{\text{delay}}$.

\begin{table}[t]
\caption{Quantitative comparison on synthetic double-regime-change epidemics with proportional zero-inflated mask in the obserations. All regression metrics are reported as Median [$Q_1, Q_3$] over 100 simulations.}
\label{tab:ablation_full}
\begin{center}
\begin{tiny}
\setlength{\tabcolsep}{3pt} 
\begin{tabular}{l rccc}
\toprule
Method  & $\Delta_{\text{delay}}$ & MDR & RMSE & MAE \\
\midrule
Full CIRL
& \underline{5.00}[2.00,9.00] & 0.02& \textbf{0.51}[0.44,0.58] & \textbf{0.36}[0.31,0.42] 
\\

w/o \emph{TCE}
& \textbf{3.00}[1.00,7.75] & 0.02 & 0.72[0.59,0.83] & \underline{0.48}[0.40,0.54]
\\

w/o \emph{HIE}
& -9.00[-10.00,-7.00] & 0.018 & \underline{0.67}[0.66,0.68] & 0.54[0.53,0.55]
\\

w/ \emph{MSE}
& -7.00[-11.25,-3.00] & 0.00 & 0.75[0.67,0.84] & 0.52[0.45,0.61]
\\

w/ \emph{Poisson}
& -10.00[-15.00,-5.00] & 0.00 & 0.78[0.65,0.88] & 0.51[0.43,0.58]
\\
\bottomrule
\end{tabular}
\end{tiny}
\end{center}
\vskip -0.1in
\end{table}

\subsection{Ablation Study}
To assess the contribution of each model component under realistic surveillance conditions, we conducted an ablation study on datasets with proportional zero-inflated mask, revealing their complementary roles in shaping the accuracy–responsiveness trade-off.

Removing the \emph{Temporal Coordinate Embedding (TCE)} module results in the lowest latency (median $\Delta_{\text{delay}}=3.00$), suggesting a highly reactive model. However, this comes at the cost of over-reacting to noise: while it achieves the second-best MAE, its RMSE deteriorates significantly to $0.72$. This discrepancy (decent MAE vs. poor RMSE) indicates that without TCE, the model suffers from sporadic large errors/spikes, confirming TCE's role in suppressing extreme outliers.

Conversely, excluding the \emph{Historical Incidence Encoder (HIE)} leads to a reduction in IQR of the delay, implying a more rigid response pattern. While this variant achieves the second-best RMSE and an MAE of $0.54$, it fails to attain the optimal precision of the full model. This suggests that HIE introduces necessary temporal flexibility, allowing the model to adaptively adjust its lag for different scenarios rather than defaulting to a fixed (low-variance) but suboptimal response.

Furthermore, we find that replacing the ZIP-based likelihood of the \emph{Renewal-based Zero-Inflated Poisson Observation Model} with standard MSE-based (using Mean Square Error loss) or Poisson-based (using Poisson likelihood) variants leads to consistent performance degradation, with RMSE rising to 0.75 and 0.78, respectively. This indicates that the ZIP likelihood functions beyond regularization, acting as a principled observation model that coherently aligns the learning objective with the sparse and zero-inflated data-generating process.
Consequently, the proposed architecture does not rely on a single module but leverages the synergy of all components to achieve a robust equilibrium between timeliness and accuracy.

\subsection{Empirical Validation on Real-World Epidemic Dynamics}
Finally, given the absence of ground truth $\mathcal{R}$ in real-world settings, we assess the performance of our framework  by focusing on epidemiological plausibility and internal consistency.
We train the CIRL model on the observed period to retrospectively infer and extrapolate $\mathcal{R}$ trajectories (10-day horizon), benchmarking our $\mathcal{R}$ estimates against retrospective baseline estimates obtained from the full time series, and assessing the alignment between model-generated incidence and reported case counts.

\begin{figure}[ht]
    \centering
    \begin{subfigure}[b]{0.48\linewidth}
        \centering
        \includegraphics[width=\linewidth]{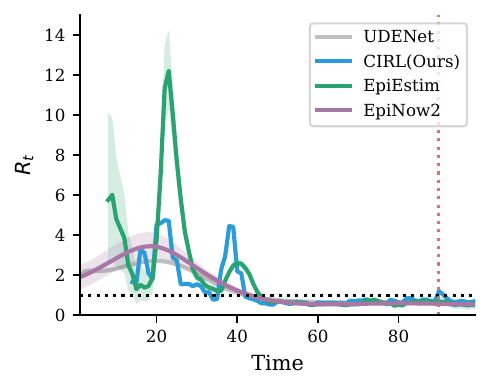}
        \caption{$R_t$ trajectory} 
        \label{fig:sars_1}
    \end{subfigure}
    \hfill 
    \begin{subfigure}[b]{0.48\linewidth}
        \centering
        \includegraphics[width=\linewidth]{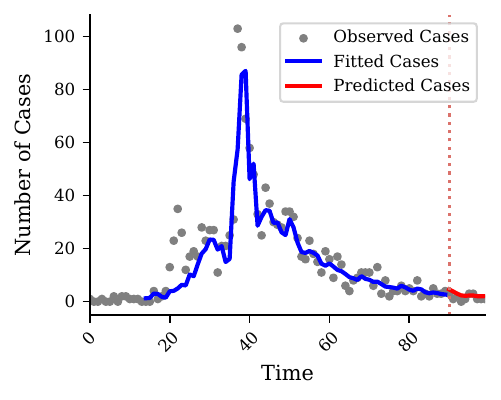}
        \caption{Fitted \& Projected incidence} 
        \label{fig:sars_2}
    \end{subfigure}

    \caption{Evaluation during the 2003 SARS in Hong Kong. (a) Retrospective $R_t$ estimation by CIRL and baseline methods, with CIRL additionally generating short-term future $R_t$ trajectories.  (b) Incidence fitting and short-horizon projection implied by CIRL. The comparison highlights the robustness of the inferred transmission dynamics to observation sparsity.
}
    \label{fig:sars_result}
    \vskip -0.1in
\end{figure}

\paragraph{SARS 2003 (Hong Kong, China).}
We utilized the canonical daily incidence dataset of the 2003 Hong Kong SARS outbreak (February–June 2003), sourced from the EpiEstim R package~\cite{cori2013new}.
Our inferred $R_t$ trajectory closely follows that of EpiEstim, as shown in Figure~\ref{fig:sars_1}, capturing the two prominent peaks identified by EpiEstim, albeit with slightly reduced fluctuations. During the late containment phase, our model's prospective forecasts consistently indicate $R_t<1.0$, in agreement with all comparison methods, reflecting the effective control of the outbreak. Furthermore, The model demonstrates high predictive consistency for incidence (Figure~\ref{fig:sars_2}), with RMSE and MAE below 2 (Appendix~\ref{app:incidence-metrics}), indicating that the inferred transmission dynamics reliably reconstruct the observed epidemic trajectory.

\begin{figure}[ht]
    \centering
    \begin{subfigure}[b]{0.48\linewidth}
        \centering
        \includegraphics[width=\linewidth]{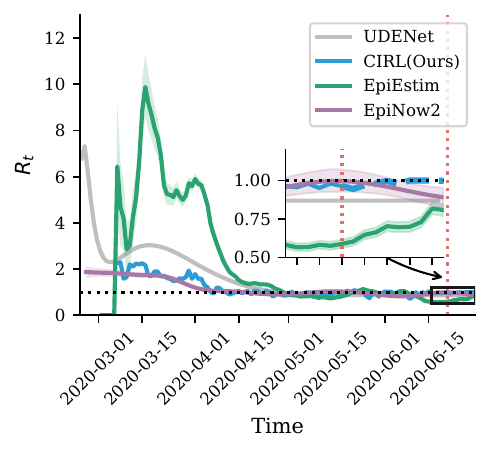}
        \caption{$R_t$ trajectory} 
        \label{fig:covid_1}
    \end{subfigure}
    \hfill 
    \begin{subfigure}[b]{0.48\linewidth}
        \centering
        \includegraphics[width=\linewidth]{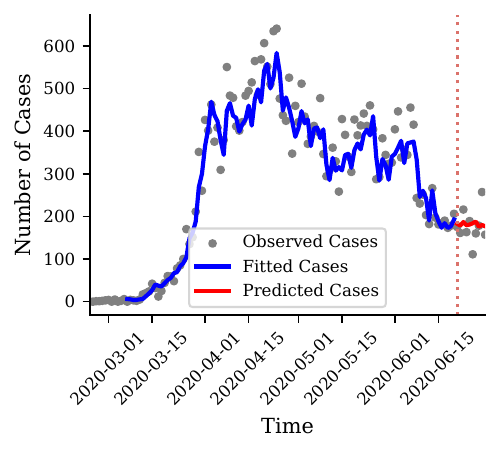}
        \caption{Fitted \& Projected incidence} 
        \label{fig:covid_2}
    \end{subfigure}

    \caption{Evaluation on the Ontario COVID-19 first wave. (a) Comparison of retrospective $R_t$ estimates. (b) Incidence fitting and implied projection. The results provide a qualitative assessment of the internal consistency of the learned conditional inverse mapping under noisy and non-stationary real-world observations.
}
    \label{fig:covid_result}
    \vskip -0.1in
\end{figure}
\paragraph{COVID-19 (Ontario, Canada).}
We analyzed the first wave of the COVID-19 epidemic in Ontario, Canada (January–June 2020), using the comparison benchmarks reported in UDENet~\cite{song2022estimating}.
As shown in Figure~\ref{fig:covid_1}, our inferred $R_t$ trajectory closely follows the overall trend of EpiNow2, while exhibiting variability comparable to that of EpiEstim but with slightly reduced fluctuation magnitude. Following epidemic control in May, all methods estimate $R_t$ to be near, yet below, the critical threshold of 1.0.
Notably, over the final ten days, our predicted $R_t$ remains close to the EpiNow2 estimates while stabilizing nearer to 1.0 toward the terminal stage, reflecting a conservative characterization of residual transmission dynamics.
Meanwhile, the predicted incidence ($I_t$) produced by our model closely matches the observed trend (Figure~\ref{fig:covid_2}), with both RMSE and MAE smaller than those obtained from direct fitting (Appendix~\ref{app:incidence-metrics}), suggesting that the learned conditional inverse mapping does not exhibit overfitting.

\section{Conclusion}
We presented a Conditional Inverse Reproduction Learning (CIRL) framework for estimating time-varying reproduction numbers from epidemic incidence data.
By formulating the task as a conditional inverse problem, CIRL learns a direct mapping from historical incidence patterns and time information to latent reproduction numbers, avoiding strong parametric assumptions or piecewise-constant constraints commonly imposed in existing approaches.
The proposed framework integrates epidemiological consistency through the renewal equation as a forward operator, while employing a Zero-Inflated Poisson observation model to account for inherent reporting noise and excess zeros ubiquitous in real-world surveillance. 
Without imposing parametric or piecewise-constant assumptions on the temporal evolution of $R_t$, the learned mapping remains responsive to abrupt transmission changes.
Evaluations across synthetic and empirical benchmarks confirm CIRL's precise $R_t$ inference, which is further corroborated by high-fidelity short-term forward simulations.
A natural extension of this work is to jointly infer the generation interval distribution within the same conditional inverse framework, further reducing reliance on external epidemiological assumptions.

\section*{Impact Statement}
This paper presents work whose goal is to advance the field of Machine Learning. There are many potential societal consequences of our work, none which we feel must be specifically highlighted here.


\bibliography{ref}
\bibliographystyle{icml2026}

\newpage

\appendix
\onecolumn 

\section{Failure Mode Analysis}
\label{app:failure_analysis}

\begin{figure}[ht]
    \centering
        \begin{subfigure}[b]{0.48\linewidth}
        \centering
        \includegraphics[width=\linewidth]{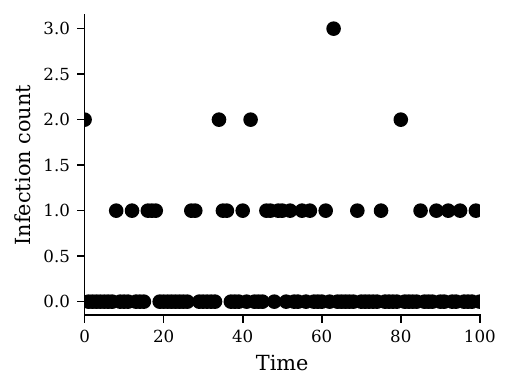}
        \caption{Infection counts} 
        \label{fig:fail1_a}
    \end{subfigure}
    \hfill 
    \begin{subfigure}[b]{0.48\linewidth}
        \centering
        \includegraphics[width=\linewidth]{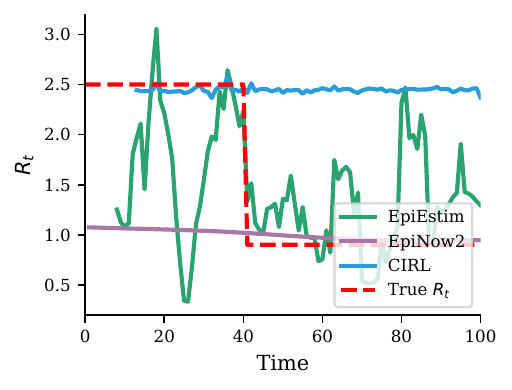}
        \caption{Time-varying reproduction number} 
        \label{fig:fail1_b}
    \end{subfigure}
    \caption{Visual analysis of the solitary failure case. (a) shows the synthetic incidence data, characterized by extreme sparsity and low counts ($y_{max} \le 4$). (b) exhibits the estimated $R_t$ trajectory. Due to the indistinguishability between the weak transmission signal and background noise, the model conservatively maintained a flat trend, resulting in a missed detection of the $R_t$ shift.}
    \label{fig:failure_case}
\end{figure}

In Section \ref{sec:experiments}, we reported a single failure instance out of 100 simulations (MDR = 0.01). To provide complete transparency, we visualized the incidenc data and model estimation for this specific instance of single-step regime shift scenario in Figure \ref{fig:failure_case}.

As illustrated in Figure~\ref{fig:failure_case}, this simulation represents an extreme edge case characterized by critical data sparsity. The maximum daily incidence count recorded is only 3, with a zero-proportion exceeding 50\%. In such a low-count regime, the signal-to-noise ratio is severely degraded, making the mathematical distinction between stochastic zeros and a true reduction in transmissibility ambiguous. Consequently, the CIRL model predicts a flat trend (blue line), failing to capture the ground-truth step change (red dashed line) at $t=40$. This indicates that in the absence of strong data evidence, the model defaults to a conservative estimate rather than "hallucinating" a regime shift.

However, a comparative analysis of baseline behaviors highlights the relative robustness of our approach even in this failure case. While CIRL misses the detection, standard methods exhibit more detrimental failure modes. EpiEstim (green line) displays extreme volatility, oscillating violently between $R_t$ values of approximately 0.5 and 3.0. This behavior confirms that standard renewal methods are prone to overfitting stochastic noise in low-incidence settings. Similarly, EpiNow2 (purple line) produces an incorrect linear drift, failing to identify the abrupt dynamics. Thus, while quantitatively categorized as a miss for CIRL, this qualitative examination suggests that our framework maintains a higher degree of safety and stability compared to baselines, avoiding noise-induced volatility when the data quality is insufficient for reliable inference.

\section{Additional Results on Incidence Fitting and Prediction}
\label{app:incidence-metrics}

In addition to the main evaluations on the reproduction number $R_t$, we report quantitative results on incidence fitting and short-term prediction on real-world datasets.
Specifically, we compute RMSE and MAE between the observed incidence and the corresponding fitted and predicted incidence $\hat{I}_t$ produced by our model, as shown in Table~\ref{tab:incidence-results}.

\begin{table}[ht] 
\caption{Evaluation of incidence reconstruction and short-term forecasting accuracy.
RMSE and MAE are reported for both in-sample fitting and rolling prospective forecasts of incidence ($I_t$).
The fitting error measures how well the inferred transmission dynamics reconstruct the observed data, while the rolling forecast error assesses the temporal consistency and generalization of the learned conditional inverse mapping.}
\label{tab:incidence-results}
\begin{center}
\begin{small}
\setlength{\tabcolsep}{4pt} 
\begin{tabular}{l cc cc}
\toprule
\multirow{2}{*}{Datasets} & \multicolumn{2}{c}{SARS 2003} & \multicolumn{2}{c}{COVID-19} \\
\cmidrule(r){2-3} \cmidrule(l){4-5}
& RMSE & MAE & RMSE & MAE \\
\midrule
Fitting & 8.62 & 5.05 & 45.34 & 33.58 \\
Prediction & 1.51 & 1.35 & 37.41 & 27.84 \\
\bottomrule
\end{tabular}
\end{small}
\end{center}
\vskip -0.1in
\end{table}

These results are provided in the appendix for completeness, as the primary focus of this work is the inference and forecasting of latent transmission dynamics rather than direct optimization for incidence-level accuracy.
Nevertheless, accurate incidence reconstruction and prediction serve as an important consistency check, ensuring that the inferred and predicted $R_t$ jointly explains both the observed and future incidence patterns.

\section{Reconstructing Latent Dynamics from Noisy Observations}
\label{app:incidence-reconstruction}
In this section, we provide a detailed analysis of the reconstructed incidence ($\hat{I}_t$), which is the intermediate outputs generated during the $R_t$ estimation process described in the main text.

\subsection{Evaluation Scenarios and Metrics}
We analyze the model outputs across the three synthetic scenarios of the main experiments:
\begin{enumerate}
    \item \textbf{Scenario-1}: A single abrupt change in transmissibility.
    \item \textbf{Scenario-2}: Double-step shifts in $R_t$.
    \item \textbf{Scenario-3}: The incidence observations of \textbf{Scenario-2} with \emph{proportional zero-inflated mask}.
\end{enumerate}

To rigorously assess the quality of the reconstructed $\hat{I}_t$, we compare it against distinct ground-truth targets derived directly from the simulation data generation process:

\begin{itemize}
    \item \textbf{$\lambda_{true}$}: The time-varying Poisson rate parameter used to generate the synthetic data~\cite{dai2023new}. This represents the theoretical expectation of daily cases derived from the true $R_t$.
    \item \textbf{$I_{obs}$}: The final synthetic data used for model training, which includes stochastic Poisson noise (\textbf{Scenario-1} and \textbf{Scenario-2}) and artificial reporting zeros (\textbf{Scenario-3}). Comparison here assesses how well the model fits the training data with structural noise.
    \item \textbf{$I_{raw}$}: Specifically for \textbf{Scenario-3}, these are the incidence counts from \textbf{Scenario-2}. Comparing against this target evaluates the model's ability to recover the original case counts lost to reporting failures. 
    \item \textbf{$\lambda_{renewal}$}: The expected incidence calculated using incidence counts revised by the standard renewal equation under the true $R_t$. Comparison against this target evaluates the {mathematical consistency} of the framework (i.e., verifying that the estimated $R_t$ and the reconstructed incidence are aligned via the epidemiological renewal process).
\end{itemize}

\subsection{Performance Analysis}
Table~\ref{tab:reconstruction_performance} and Table~\ref{tab:diagnostics} summarize the quantitative results. Below, we discuss the implications of these metrics regarding latent trend recovery, overfitting risks, and robustness to reporting failures. 

\begin{table}[h]
    \centering
    \caption{Reconstruction Performance. RMSE of estimated incidence $\hat{I}_t$ against the true Poisson rate $\lambda_{true}$. Values are reported as the Median $[Q_1, Q_3]$.}
    \label{tab:reconstruction_performance}
    \setlength{\tabcolsep}{3pt}
    \begin{small}
    \begin{tabular}{l c c c c}
        \toprule
        \textbf{Dataset Scenario} & \textbf{Comparison Target} & \textbf{CIRL(Ours)} & \textbf{EpiNow2} & \textbf{EpiEstim} \\
        \midrule
        \textbf{Scenario-1} & $\lambda_{true}$ & \underline{10.04}[~6.03,14.10] & ~\textbf{8.45}[~5.21,12.03] & 10.26[~6.31,14.06]\\
        \textbf{Scenario-2} & $\lambda_{true}$ &  ~7.13[~4.66,10.81] & ~\textbf{5.93}[~4.12,~8.45] & ~\underline{7.03}[~4.88,10.44] \\
        \textbf{Scenario-3} & $\lambda_{true}$ &  ~\underline{7.04}[~4.86,10.39] & ~\textbf{6.80}[~4.49,10.66] & ~7.83[~5.34,12.08] \\
        \bottomrule
    \end{tabular}
    \end{small}
\end{table}

\begin{table}[h]
    \centering
    \caption{{Reconstruction Metrics.} Values are Median [$Q_1, Q_3$]. 
    Left (Fitting to Noisy Observations):  Near-zero error on $I_{obs}$ indicates overfitting to noise. 
    Middle (Internal Consistency): Low error against $\lambda_{renewal}$ confirms mathematical validity. 
    Right (Reconstruction / Counterfactual Recovery): Evaluated solely on \textbf{Scenario-3}, measuring the error against the \emph{unmasked truth} ($I_{raw}$). Note that while baselines fit $I_{obs}$ well, they fail to recover the hidden cases.}
    \label{tab:diagnostics}
    \setlength{\tabcolsep}{3pt} 
    \begin{small}
    \begin{tabular}{l  ccc  ccc  c}
        \toprule
        \multirow{3}{*}{\textbf{Method}} & \multicolumn{3}{c}{\textbf{Fitting to Noisy Observations}} & \multicolumn{3}{c}{\textbf{Internal Consistency}} & \textbf{Reconstruction} \\
        & \multicolumn{3}{c}{(Target: $I_{obs}$)} & \multicolumn{3}{c}{(Target: $\lambda_{renewal}$)} & (Target: $I_{raw}$) \\
        \cmidrule(lr){2-4} \cmidrule(lr){5-7} \cmidrule(lr){8-8}
         & \textbf{Scenario-1} & \textbf{Scenario-2} & \textbf{Scenario-3} & \textbf{Scenario-1} & \textbf{Scenario-2} & \textbf{Scenario-3} & \textbf{Scenario-3} \\
        \midrule
        EpiEstim & \textbf{0.00}[0.00,0.00] & \textbf{0.00}[0.00,0.00] & \textbf{0.00}[0.00,0.00] & 
        4.41[3.37,5.87] & 3.90[3.01,5.31] & 4.76[3.56,7.50] & \underline{3.16}[1.90,5.19] \\
        EpiNow2  & 3.91[2.87,4.79] & 3.40[2.71,4.66] & 4.53[3.26,7.17] & 
        \textbf{3.31}[2.11,4.51] & \textbf{2.78}[1.99,3.75] & \underline{3.59}[2.35,5.77] & 4.01[3.05,6.02] \\
        CIRL(Ours)& \underline{1.94}[1.55,2.42] & \underline{1.89}[1.58,2.47] & \underline{3.67}[2.51, 5.86] & 
        \underline{3.62}[2.66,5.35] & \underline{3.31}[2.35,4.85] & \textbf{3.03}[2.23,4.32] & \textbf{2.44}[1.91,3.06] \\
        \bottomrule
    \end{tabular}
    \end{small}
\end{table}

\paragraph{Recovery of Latent Dynamics and Generalization.} 
As detailed in Table~\ref{tab:reconstruction_performance}, the Gaussian Process-based baseline, EpiNow2, exhibits strong performance in standard scenarios, achieving the lowest RMSE against the true latent intensity $\lambda_{true}$ due to its smoothness priors. Our proposed CIRL framework demonstrates competitive stability. In Scenario-2, CIRL achieves a median RMSE of 7.13, which is statistically indistinguishable from EpiNow2's 5.93 and EpiEstim's 7.03 given the substantial overlap in interquartile intervals. 
However, the key distinction lies in consistency. While EpiEstim performs comparably in Scenario-2, it yields higher errors in Scenario-1 (10.26 vs. 10.04) and Scenario-3 (7.83 vs. 7.04). This fluctuation indicates that EpiEstim's reliance on observations makes it sensitive to stochastic variations, whereas CIRL maintains a more robust performance profile across varying degrees of complexity. 

\paragraph{The Overfitting Trap versus Rigid Underfitting.}
A critical diagnostic insight is provided by the evaluation against noisy observations ($I_{obs}$) in Table~\ref{tab:diagnostics} (Left). EpiEstim achieves a median RMSE of 0.00 across all scenarios. While mathematically perfect, this zero-error fit reveals severe overfitting, confirming that the model is tracking stochastic Poisson noise and reporting artifacts rather than the epidemiological trend. 
In contrast, EpiNow2 consistently exhibits the highest RMSE against observations across all scenarios. While this indicates a resistance to noise, the elevated error in Scenario-3 (4.53 compared to CIRL's 3.67) suggests that its smoothing mechanism becomes {overly rigid} in the presence of zero-inflated data, leading to underfitting. 
CIRL strikes a balance with RMSE values ranging from 1.89 to 3.67, fitting the data sufficiently well to learn the trend while maintaining enough deviation to avoid modeling the noise.

\paragraph{Robustness to Sparsity and Counterfactual Reconstruction.}
The advantages of the proposed framework become most pronounced in Scenario-3, where reporting failures introduce structural zeros. As shown in Table~\ref{tab:diagnostics} (Middle), while EpiNow2 performs well in standard scenarios (Scenario-1 and Scenario-2), its internal consistency deteriorates in Scenario-3 (RMSE 3.59). In contrast, CIRL achieves the lowest consistency error (RMSE 3.03) in this challenging setting, confirming that its inferred $R_t$ remains mathematically aligned with the reconstructed incidence even under severe data sparsity.

Most importantly, the Counterfactual Reconstruction metric (Table~\ref{tab:diagnostics}, Right) provides the definitive evidence of our method's efficacy. When evaluated against the unmasked truth ($I_{raw}$), EpiNow2 yields a high RMSE of 4.01, implying that its smoothing priors are biased downwards by the artificial zeros. CIRL, however, achieves a significantly lower RMSE of 2.44. By explicitly modeling the zero-inflation probability ($\pi_t$), CIRL effectively identifies structural reporting failures and "fills in" the missing cases, thereby recovering the {true underlying incidence} that baselines fail to capture.

Most importantly, the counterfactual recovery (reconstruction) metric (Table~\ref{tab:diagnostics}, right) offers the most direct evidence of the effectiveness of our method. When evaluated against the unmasked truth ($I_{raw}$), EpiNow2 yields a high RMSE of 4.01, implying that its smoothing priors are biased by the artificial zeros. CIRL, however, achieves a significantly lower RMSE of 2.44. By explicitly modeling the zero-inflation probability ($\pi_t$), CIRL effectively identifies structural reporting failures and "fills in" the missing cases, thereby recovering the {true underlying incidence} that baselines fail to capture.

\end{document}